\documentclass{article}

\usepackage{microtype}
\usepackage{graphicx}
\usepackage{subcaption} 
\usepackage{booktabs} 

\usepackage{multirow}
\usepackage[table]{xcolor}  
\newcommand{\hl}[1]{\cellcolor[HTML]{F2F2F2}#1}

\usepackage[linesnumbered,ruled,vlined,algo2e]{algorithm2e}
\usepackage{makecell}

\usepackage{hyperref}

\usepackage{amsmath,amssymb,amsthm}

\providecommand{\ff}{\mathbf{f}}

\providecommand{\ww}{\mathbf{w}}
\providecommand{\xx}{\mathbf{x}}

\providecommand{\cD}{\mathcal{D}}

\providecommand{\cO}{\mathcal{O}}

\newenvironment{talign*}
{\csname align*\endcsname}
{\endalign}

\usepackage[utf8]{inputenc}         %
\usepackage[T1]{fontenc}            %
\usepackage{url}                    %
\usepackage{booktabs}               %
\usepackage{amsfonts}               %
\usepackage{nicefrac}               %
\usepackage{microtype}              %
\usepackage{xcolor}                 %
\usepackage{graphicx}
\usepackage{subcaption}
\usepackage[flushleft]{threeparttable}
\usepackage{float}
\usepackage{multirow}
\usepackage{xspace}
\usepackage{natbib}
\usepackage{enumitem}
\usepackage[font=small]{caption}
\usepackage{autobreak}
\usepackage{sidecap}
\usepackage{wrapfig}
\usepackage{bbding}
\usepackage[toc, page, header]{appendix}
\usepackage{tikz}
\usepackage{xcolor}
\usepackage{pifont}
\usepackage{mdframed}
\usepackage{colortbl}

\hypersetup{
  colorlinks=true,
  linkcolor=blue,
  citecolor=blue,
  urlcolor=blue
}

\definecolor{coral}{RGB}{255,127,80}
\definecolor{darkgreen}{RGB}{0,100,0}
\definecolor{darkyellow}{RGB}{204,153,0}
\definecolor{salmon}{RGB}{250,128,114}
\definecolor{darkred}{RGB}{150,0,0}
\newcommand{\darkredtext}[1]{{\color{darkred}#1}}

\newcommand{\algopt}{\textsc{\texttt{GoldDiff}}\xspace}

\newcommand{\secref}[1]{\hyperref[#1]{\darkredtext{Sec.~\ref*{#1}}}}
\newcommand{\thmref}[1]{\hyperref[#1]{\darkredtext{Thm.~\ref*{#1}}}}
\newcommand{\defref}[1]{\hyperref[#1]{\darkredtext{Def.~\ref*{#1}}}}
\newcommand{\propref}[1]{\hyperref[#1]{\darkredtext{Prop.~\ref*{#1}}}}
\newcommand{\assumpref}[1]{\hyperref[#1]{\darkredtext{Assump.~\ref*{#1}}}}
\newcommand{\remarkref}[1]{\hyperref[#1]{\darkredtext{Rem.~\ref*{#1}}}}
\newcommand{\hypref}[1]{\hyperref[#1]{\darkredtext{Hyp.~\ref*{#1}}}}
\newcommand{\conjref}[1]{\hyperref[#1]{\darkredtext{Conj.~\ref*{#1}}}}
\newcommand{\lemref}[1]{\hyperref[#1]{\darkredtext{Lem.~\ref*{#1}}}}
\newcommand{\corref}[1]{\hyperref[#1]{\darkredtext{Cor.~\ref*{#1}}}}
\newcommand{\noteref}[1]{\hyperref[#1]{\darkredtext{Nota.~\ref*{#1}}}}
\newcommand{\claimref}[1]{\hyperref[#1]{\darkredtext{Clm.~\ref*{#1}}}}
\newcommand{\obsref}[1]{\hyperref[#1]{\darkredtext{Obs.~\ref*{#1}}}}
\newcommand{\algref}[1]{\hyperref[#1]{\darkredtext{Alg.~\ref*{#1}}}}
\newcommand{\figref}[1]{\hyperref[#1]{\darkredtext{Fig.~\ref*{#1}}}}
\newcommand{\tabref}[1]{\hyperref[#1]{\darkredtext{Tab.~\ref*{#1}}}}
\newcommand{\appref}[1]{\hyperref[#1]{\darkredtext{App.~\ref*{#1}}}}

\newtheoremstyle{custom}
{1pt} %
{1pt} %
{\itshape} %
{} %
{\bfseries} %
{} %
{ } %
{\thmname{#1} \thmnumber{#2} \thmnote{(#3)} . } %

\theoremstyle{custom}

\newtheorem{innerdefinition}{Definition}
\newtheorem{innerproposition}{Proposition}
\newtheorem{innerassumption}{Assumption}
\newtheorem{innerremark}{Remark}
\newtheorem{innertheorem}{Theorem}
\newtheorem{innerhypothesis}{Hypothesis}
\newtheorem{innerconjecture}{Conjecture}
\newtheorem{innerlemma}{Lemma}
\newtheorem{innercorollary}{Corollary}
\newtheorem{innernotation}{Notation}
\newtheorem{innerclaim}{Claim}
\newtheorem{innerproblem}{Problem}

\newtheorem{innerobservation}{Observation}

\newmdenv[
  backgroundcolor=gray!10,
  linecolor=gray!100,
  linewidth=0.8pt,
  skipabove=2pt,
  skipbelow=2pt,
  innertopmargin=5pt,
  innerbottommargin=5pt,
  innerleftmargin=5pt,
  innerrightmargin=5pt,
]{definitionframe}

\newmdenv[
  backgroundcolor=blue!10,
  linecolor=blue!100,
  linewidth=0.8pt,
  skipabove=2pt,
  skipbelow=2pt,
  innertopmargin=5pt,
  innerbottommargin=5pt,
  innerleftmargin=5pt,
  innerrightmargin=5pt,
]{propositionframe}

\newmdenv[
  backgroundcolor=green!10,
  linecolor=green!100,
  linewidth=0.8pt,
  skipabove=2pt,
  skipbelow=2pt,
  innertopmargin=5pt,
  innerbottommargin=5pt,
  innerleftmargin=5pt,
  innerrightmargin=5pt,
]{assumptionframe}

\newmdenv[
  backgroundcolor=yellow!10,
  linecolor=yellow!100,
  linewidth=0.8pt,
  skipabove=2pt,
  skipbelow=2pt,
  innertopmargin=5pt,
  innerbottommargin=5pt,
  innerleftmargin=5pt,
  innerrightmargin=5pt,
]{remarkframe}

\newmdenv[
  backgroundcolor=cyan!5,
  linecolor=cyan!30,
  linewidth=0.8pt,
  skipabove=2pt,
  skipbelow=2pt,
  innertopmargin=10pt,
  innerbottommargin=5pt,
  innerleftmargin=5pt,
  innerrightmargin=5pt,
]{theoremframe}

\newmdenv[
  backgroundcolor=purple!10,
  linecolor=purple!100,
  linewidth=0.8pt,
  skipabove=2pt,
  skipbelow=2pt,
  innertopmargin=5pt,
  innerbottommargin=5pt,
  innerleftmargin=5pt,
  innerrightmargin=5pt,
]{hypothesisframe}

\newmdenv[
  backgroundcolor=orange!10,
  linecolor=orange!100,
  linewidth=0.8pt,
  skipabove=2pt,
  skipbelow=2pt,
  innertopmargin=5pt,
  innerbottommargin=5pt,
  innerleftmargin=5pt,
  innerrightmargin=5pt,
]{conjectureframe}

\newmdenv[
  backgroundcolor=cyan!10,
  linecolor=cyan!100,
  linewidth=0.8pt,
  skipabove=2pt,
  skipbelow=2pt,
  innertopmargin=10pt,
  innerbottommargin=5pt,
  innerleftmargin=5pt,
  innerrightmargin=5pt,
]{lemmaframe}

\newmdenv[
  backgroundcolor=magenta!5,
  linecolor=magenta!30,
  linewidth=0.8pt,
  skipabove=2pt,
  skipbelow=2pt,
  innertopmargin=10pt,
  innerbottommargin=5pt,
  innerleftmargin=5pt,
  innerrightmargin=5pt,
]{corollaryframe}

\newmdenv[
  backgroundcolor=pink!10,
  linecolor=pink!100,
  linewidth=0.8pt,
  skipabove=2pt,
  skipbelow=2pt,
  innertopmargin=5pt,
  innerbottommargin=5pt,
  innerleftmargin=5pt,
  innerrightmargin=5pt,
]{notationframe}

\newmdenv[
  backgroundcolor=violet!10,
  linecolor=violet!100,
  linewidth=0.8pt,
  skipabove=2pt,
  skipbelow=2pt,
  innertopmargin=5pt,
  innerbottommargin=5pt,
  innerleftmargin=5pt,
  innerrightmargin=5pt,
]{claimframe}

\newmdenv[
  backgroundcolor=salmon!10,
  linecolor=salmon!100,
  linewidth=0.8pt,
  skipabove=2pt,
  skipbelow=2pt,
  innertopmargin=5pt,
  innerbottommargin=5pt,
  innerleftmargin=5pt,
  innerrightmargin=5pt,
]{problemframe}

\newmdenv[
  backgroundcolor=lavender!10,
  linecolor=lavender!100,
  linewidth=0.8pt,
  skipabove=2pt,
  skipbelow=2pt,
  innertopmargin=5pt,
  innerbottommargin=5pt,
  innerleftmargin=5pt,
  innerrightmargin=5pt,
]{observationframe}

\newenvironment{theorem}
{\begin{theoremframe}\begin{innertheorem}}
      {\end{innertheorem}\end{theoremframe}}

\newenvironment{corollary}
{\begin{corollaryframe}\begin{innercorollary}}
      {\end{innercorollary}\end{corollaryframe}}

\usepackage[textwidth=2.6cm,textsize=tiny]{todonotes}

\usepackage[preprint]{configuration/icml2026}

\usepackage{amsmath}
\usepackage{amssymb}
\usepackage{mathtools}
\usepackage{amsthm}

\usepackage{tikz}     
\usetikzlibrary{tikzmark, arrows.meta, calc} 

\theoremstyle{plain}

\usepackage[textsize=tiny]{todonotes}

\icmltitlerunning{Fast and Scalable Analytical Diffusion}

\begin{document}

\twocolumn[
  \icmltitle{Fast and Scalable Analytical Diffusion}



  \icmlsetsymbol{equal}{*}

  \begin{icmlauthorlist}
    \icmlauthor{Xinyi Shang}{equal,yyy1,yyy5}
    \icmlauthor{Peng Sun}{equal,yyy2,yyy3}
    \icmlauthor{Jingyu Lin}{yyy4}
    \icmlauthor{Zhiqiang Shen}{yyy5}
  \end{icmlauthorlist}

  \icmlaffiliation{yyy1}{University College London}
  \icmlaffiliation{yyy2}{Zhejiang University}
  \icmlaffiliation{yyy3}{Westlake University}
  \icmlaffiliation{yyy4}{Monash University}
  \icmlaffiliation{yyy5}{Mohamed bin Zayed University of Artificial Intelligence}

  \icmlcorrespondingauthor{Zhiqiang Shen}{Zhiqiang Shen@mbzuai.ac.ae}

  \icmlkeywords{Machine Learning, ICML}

  \vskip 0.3in
]



\printAffiliationsAndNotice{\icmlEqualContribution}  

\begin{abstract}
Analytical diffusion models offer a mathematically transparent path to generative modeling by formulating the denoising score as an empirical-Bayes posterior mean. However, this interpretability comes at a prohibitive cost: the standard formulation necessitates a full-dataset scan at every timestep, scaling linearly with dataset size.
In this work, we present the first systematic study addressing this scalability bottleneck.
We challenge the prevailing assumption that the entire training data is necessary, uncovering the phenomenon of \textit{Posterior Progressive Concentration}: the effective golden support of the denoising score is not static but shrinks asymptotically from the global manifold to a local neighborhood as the signal-to-noise ratio increases.
Capitalizing on this, we propose \textit{Dynamic Time-Aware Golden Subset Diffusion} (\algopt), a training-free framework that decouples inference complexity from dataset size.
Instead of static retrieval, \algopt uses a coarse-to-fine mechanism to dynamically pinpoint the ``Golden Subset'' for inference.
Theoretically, we derive rigorous bounds guaranteeing that our sparse approximation converges to the exact score.
Empirically, \algopt achieves a $\bf 71 \times$ speedup on AFHQ while matching or achieving even better performance than full-scan baselines.
Most notably, we demonstrate the first successful scaling of analytical diffusion to ImageNet-1K, unlocking a scalable, training-free paradigm for large-scale generative modeling.
\end{abstract}

\section{Introduction}

Analytical diffusion models~\cite{kamb2024analytic} have emerged as a principled framework for demystifying generative dynamics, offering closed-form estimators that provide mechanistic insight into denoising \cite{de2022convergence,scarvelis2023closed,kamb2024analytic}. A core primitive is the \emph{empirical Bayes denoiser}: at each diffusion step, the model forms a posterior over the training set and uses the posterior mean as the denoising score \cite{kamb2024analytic,Lukoianov2025Locality,niedoba2024towards}. This transparency, however, comes with a prohibitive computational burden: the posterior is supported on all $N$ training points, so naive inference requires a full-dataset scan per timestep, yielding $\mathcal{O}(ND)$ complexity for dataset size $N$ and dimension $D$. As a result, analytical diffusion becomes impractical at scale, and unlike the extensive work on accelerating neural diffusion solvers \cite{blattmann2022retrieval,sheynin2023knn,niedoba2024nearest}, optimizing exact analytical estimators remains largely unexplored.

\begin{figure*}[!t]
    \centering
          \includegraphics[width=\linewidth]{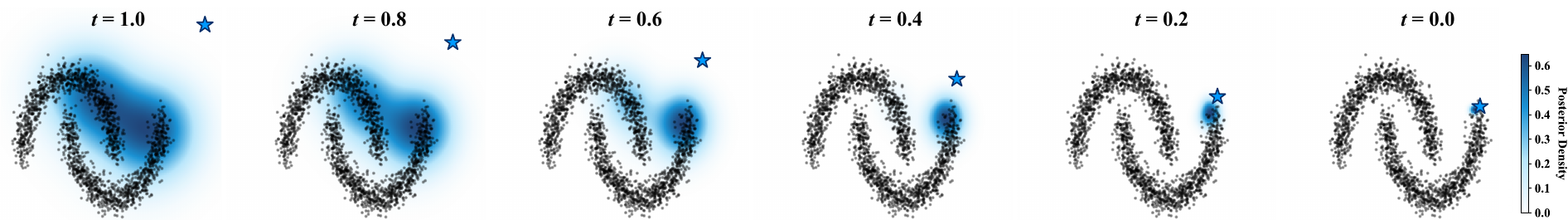}
    \caption{\textbf{The Phenomenon of Posterior Progressive Concentration.}
    \textcolor[HTML]{0099FF}{$\bigstar$} denotes the initial noise, and the target distribution is the Moons data \cite{pedregosa2011scikit}.
    As the diffusion process reverses from pure noise to data (Left to Right), the golden support of the posterior distribution \textbf{\textit{dynamically}} shrinks from the global manifold to a localized neighborhood.}
    \label{fig:motivation}
\end{figure*}

Beyond cost, scanning the entire dataset is often \emph{unnecessary} and can even be \emph{harmful}. In practice, only a small fraction of samples meaningfully contribute to the posterior at a given noise level (\figref{fig:motivation}), while irrelevant or ambiguous points may introduce bias and statistical noise, degrading the estimate. This motivates our key idea: instead of treating all training points as equally relevant, we seek a minimal, high-quality support set that preserves the empirical Bayes score while avoiding spurious influence. We formalize this as a \emph{golden subset}: a dynamically selected subset of training examples from a theoretical perspective that concentrates the posterior mass on the current noisy input, enabling both efficient inference and more reliable estimation.

In this work, we present the first systematic study addressing the acceleration of analytical diffusion models.
We challenge the conventional assumption in analytical diffusion that accurate posterior estimation requires the entire training corpus.
By systematically analyzing the spatio-temporal dynamics of posterior weights, we characterize a fundamental phenomenon termed \textit{Posterior Progressive Concentration}, where the \textit{golden support} (the subset of samples contributing non-negligible probability mass) evolves dramatically with the signal-to-noise ratio (\figref{fig:motivation}).
Specifically, in the high-noise regime, the posterior is diffuse and requires a broad support to capture global manifold structure, meaning a small static $k$ introduces severe bias (further demonstrated in \secref{sec:dynamic_method}).
Conversely, in the low-noise regime, the posterior collapses into a tight neighborhood, rendering a large static $k$ computationally wasteful without yielding accuracy gains.
This progressive concentration motivates us to decouple computational complexity from dataset size by dynamically adapting the golden support set, thereby accelerating the process without sacrificing estimation accuracy.

Building on this insight, we propose \textit{Dynamic Time-Aware Golden Subset Diffusion} (\algopt).
Departing from rigid static $k$-NN selection, \algopt employs a coarse-to-fine mechanism that dynamically pinpoints a minimal yet sufficient golden set required for accurate posterior estimation at the whole denoise time.
Crucially, our method offers benefits beyond mere speed.
We identify that existing state-of-the-art analytical models suffer from severe smoothing bias stemming from biased weight estimation, producing blurry outputs even with sufficient denoising steps (\figref{fig:wss_ss}).
By selectively the {\em Golden Subset}, we demonstrate that a simple, unbiased estimator is sufficient to unlock superior quality.
Most notably, our efficiency gains unlock a milestone for the field: we successfully scale analytical diffusion to ImageNet-1K for the first time.
\algopt achieves strong performance with a substantial acceleration, effectively bridging the gap between closed-form mathematical transparency and high-fidelity, industrial-scale generation.
\textit{Our key contributions are threefold:}

\begin{itemize}[leftmargin=1.2em]
    \item We provide the first systematic investigation into the acceleration of analytical denoisers.
    We identify the phenomenon of \textit{posterior progressive concentration}, proving that the golden support of the denoising score shrinks asymptotically as the noise level decreases.
    This finding has profound implications not only for analytical inference but also for the design of efficient neural denoisers.
    \item We propose \algopt, a training-free framework that dynamically retrieves a ``Golden Subset'' for inference.
    We derive rigorous theoretical bounds guaranteeing that our sparse approximation converges to the exact analytical score with controlled error.
    \item Extensive experiments demonstrate that the proposed \algopt achieves a {$\bf 71 \times$ \bf speedup} on AFHQ while matching or exceeding full-scan performance. Furthermore, we achieve the first successful scaling of analytical diffusion to large-scale ImageNet-1K, unlocking a new paradigm for scalable, interpretable generative modeling.
\end{itemize}

\section{Related Work}

\paragraph{Denoising Diffusion Models.}
Diffusion probabilistic models \cite{ho2020denoising,song2020score,song2020denoising} have emerged as the dominant paradigm in generative modeling, achieving state-of-the-art sample quality on a wide range of benchmarks \cite{dhariwal2021diffusion}.
Fundamentally, these models reverse a stochastic process that degrades the data distribution into noise.
Notably, when the training set is finite, the perturbed data distribution at any timestep can be characterized as a time-dependent mixture of Gaussians.
These models typically parameterize the score function $\nabla_{\xx}\log p_t(\xx)$ using deep neural networks, commonly with U-Net backbones \cite{karras2022elucidating,karras2024analyzing} or Transformer-based architectures \cite{peebles2023scalable,yao2025reconstruction}, trained via denoising score matching.
While highly effective, such learned denoisers are often treated as ``black boxes'', making it difficult to interpret or mechanistically analyze the generation dynamics \cite{kamb2024analytic}.

\noindent{\bf Analytical Diffusion.}
Analytical Diffusion models have emerged as a rigorous framework for interpreting the generative mechanisms of diffusion models \cite{kamb2024analytic}.
By formulating the score function as an empirical Bayes estimator, these methods enable exact score computation without black-box optimization.
Seminal works \cite{de2022convergence,scarvelis2023closed} establish the optimal denoiser as a theoretical proxy for understanding the behavior of deep diffusion models.
However, a central paradox lies in their generalization capabilities: the exact optimal denoiser tends to inherently memorize, collapsing to a mixture of delta functions at the training data points in the low-noise limit \cite{kamb2024analytic}.
To explain the generalization observed in neural approximations, \citet{kamb2024analytic} and \citet{niedoba2024towards} attribute it to the inductive bias (locality) of the neural architectures.
Conversely, \citet{Lukoianov2025Locality} recently posit that such locality emerges as an intrinsic statistical property of the training dataset.
Leveraging this insight, they incorporate their analytically-computed locality into the optimal denoiser-based model, demonstrating superior performance over standard analytical baselines.

\noindent{\bf Efficient Inference and Retrieval-Augmented Diffusion.}
Accelerating diffusion inference is a longstanding challenge.
While standard techniques focus on reducing the number of timesteps (e.g., distillation \cite{salimans2022progressive,yin2024improved,yin2024one}), a parallel line of research aims to reduce the per-step cost by approximating the data distribution via coresets or varying clusters \cite{niedoba2024nearest}.
In the context of retrieval-based methods, static $k$-Nearest Neighbors ($k$-NN) serves as a prevalent approximation strategy \cite{niedoba2024nearest,sheynin2023knn}.
For example, KNN-Diffusion \cite{sheynin2023knn} and RetrievalAugmented Diffusion Models \cite{blattmann2022retrieval} both condition diffusion models on the fixed $k$ nearest CLIP \cite{radford2021learning} embeddings of clean training images.

Our work diverges from these approaches in \textit{two} fundamental aspects.
First, we pioneer the optimization of analytical estimators, bridging the gap between mathematical transparency and computational scalability.
Second, we challenge the static retrieval paradigm and offer a theory-grounded time-varying posterior estimation framework.

\vspace{-5pt}
\section{Time-Aware Golden Subset Diffusion} \label{sec:method}

In this section, we present \textbf{\algopt}, a training-free acceleration framework for analytical diffusion.
To effectively retrieve the ``Golden Subset,'' we first revisit the analytical denoisers (\secref{sec:preliminary}) and explore the impact of biased weight estimation (\secref{sec:revisit_weight}).
Then, we analyze the sensitivity to subset selection, revealing a distinct \textit{two-regime behavior} governed by the signal-to-noise ratio (\secref{sec:method_sensitive}).
Motivated by these insights, we propose a theoretical-grounded dynamic selection strategy that optimizes the trade-off between retrieval recall and the aggregation budget via time-aware schedules (\secref{sec:dynamic_method}).
Furthermore, we establish rigorous theoretical guarantees to validate our dynamic mechanism and provide complexity analysis (\secref{sec:complexity}).

\vspace{-0.1in}
\subsection{Revisiting Analytical Denoisers} \label{sec:preliminary}

We define the forward diffusion process as $\xx_t = \sqrt{\alpha_t}\xx_0 + \sqrt{1-\alpha_t}\epsilon$, where $\alpha_t$ follows a monotonic signal schedule and $\epsilon \sim \mathcal{N}(\mathbf{0}, \mathbf{I})$.
The standard MSE training objective is minimized by the conditional expectation of the clean data \cite{vincent2011connection,ho2020denoising}:
\begin{equation}\small
    \min_{\ff} \mathbb{E} \left[ \| \ff(\xx_t, t) - \xx_0 \|^2 \right] \implies \hat{\ff}(\xx, t) = \mathbb{E}[\xx_0 \mid \xx_t = \xx].
    \label{eq:optimal_mmse}
\end{equation}

\noindent{\bf Optimal Empirical Bayes Denoising.}
By modeling the data prior as the empirical distribution of the training set $\cD=\{\xx_i\}_{i=1}^N$, i.e., $p(\xx_0) = \frac{1}{N}\sum_i \delta(\xx_0 - \xx_i)$, the exact \textit{Empirical Bayes Denoiser} admits a closed-form solution as a kernel-weighted average \cite{kamb2024analytic,de2022convergence,Lukoianov2025Locality}
:
\begin{equation}
    \hat{\ff}(\xx, t) = \sum_{i=1}^N \underbrace{\mathrm{softmax}_i \left( -\frac{\| \xx/\sqrt{\alpha_t} - \xx_i \|^2}{2\sigma_t^2} \right)}_{\ww_i(\xx,t)} \cdot \xx_i,
    \label{eq:empirical_denoiser}
\end{equation}
where $\sigma_t^2 \coloneqq (1-\alpha_t)/\alpha_t$ is the noise-to-signal ratio. 
Eq. \eqref{eq:empirical_denoiser} interprets the optimal denoiser as a \emph{distance-weighted average over the global training set}.

\begin{figure}[!t]\small
    \centering
    \includegraphics[width=\linewidth]{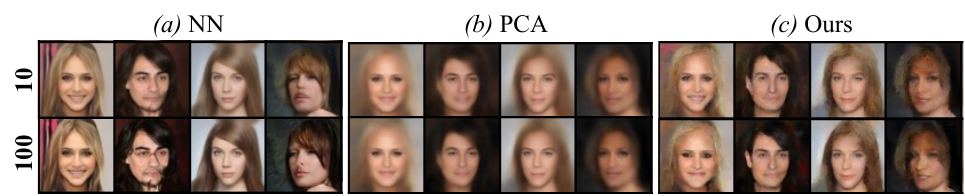}
   \caption{\textbf{Impact of Biased Weight Estimation.}
   Due to biased weight estimation, PCA \cite{Lukoianov2025Locality} produces inherently over-smoothed outputs even with sufficient denoising steps.}
    \label{fig:wss_ss}
    \vspace{-15pt}
\end{figure}

\begin{figure*}[!t]
    \centering
     \begin{subfigure}[t]{0.32\textwidth}
        \centering
        \includegraphics[width=\linewidth]{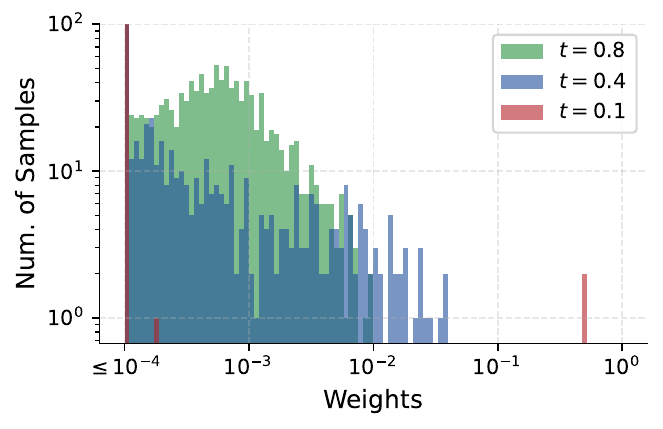}
        \caption{Posterior Weights Distribution}
        \label{fig:weight_dist}
    \end{subfigure}
    \begin{subfigure}[t]{0.32\textwidth}
        \centering
        \includegraphics[width=\linewidth]{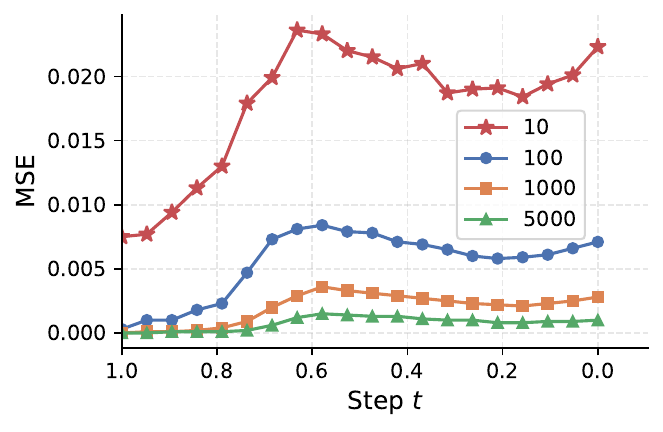}
        \caption{MSE between Subset and Full}
        \label{fig:mse_full_subset}
    \end{subfigure}
    \hfill
    \begin{subfigure}[t]{0.32\textwidth}
        \centering
        \includegraphics[width=\linewidth]{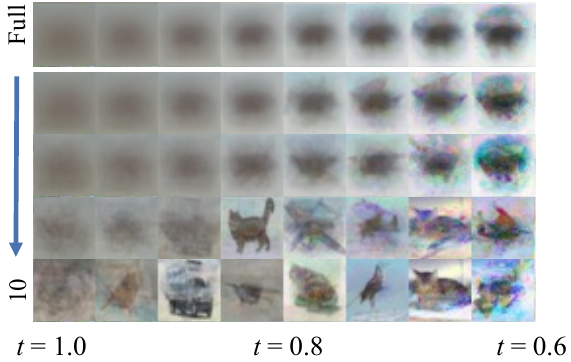}
        \caption{Qualitative Comparison}
        \label{fig:vis_full_subset}
    \end{subfigure}
    \caption{\textbf{Analysis of the SOTA method PCA \cite{Lukoianov2025Locality} for Subset Selection.} 
    (a) Evolution of the weight distribution during the denoising process. 
     (b)--(c) Sensitivity analysis: (b) Performance evaluation of the analytical denoiser across varying random subset sizes ($N_{\text{sub}} \in \{10, 100, 1000, 5000\}$) compared to the full CIFAR-10 dataset. 
     (c) Visualization of intermediate generation outputs. 
    }
\label{fig:pca_analysis}
\vspace{-10pt}
\end{figure*}

\noindent{\bf Manifold Locality and Generalization.}
While Eq. \eqref{eq:empirical_denoiser} is optimal in the MMSE sense, \citet{kamb2024analytic} reveals that it suffers from severe memorization, tending to reproduce training samples rather than generating novel content.
They reveal that the generalization ability of diffusion models arises from \textit{locality}, proposing a patch-based approximation that restricts attention to local spatial neighborhoods.
Building on this, \citet{Lukoianov2025Locality} further demonstrates that locality is an intrinsic statistical property of the image manifold. 
They propose projecting data onto local Principal Component Analysis (PCA) bases to better capture this manifold structure.
Generalizing these local variants, the denoising score can be expressed as:
\begin{equation}
    \setlength{\abovedisplayskip}{4pt}
    \setlength{\belowdisplayskip}{4pt}
    \hat{\ff}(\xx_t, t) = \sum_{i=1}^N \tilde{\ww}_i(\xx_t) \cdot \mathcal{P}_i(\xx_i),
    \label{eq:unified_local_denoiser}
\end{equation}
where $\mathcal{P}_i$ denotes a generalized local operator (e.g., patch extraction in \cite{kamb2024analytic} or PCA projection in \cite{Lukoianov2025Locality}), and $\tilde{\ww}_i$ represents the weights computed within that local subspace.
Crucially, regardless of the specific projection $\mathcal{P}_i$, the computational bottleneck remains the summation over $N$ training samples, which scales linearly as $\cO(ND)$, rendering exact inference intractable for large-scale datasets.

\vspace{-0.1in}
\subsection{Unbiased Weight Estimation}\label{sec:revisit_weight}
The current state-of-the-art PCA method \cite{Lukoianov2025Locality} computes posterior weights over the {\em entire} training set. In practice, the dataset is highly heterogeneous: some samples are truly informative, others are largely irrelevant, and a nontrivial portion can even be harmful (e.g., noisy or misleading neighbors).
This global weighting often produces an overly sharp, heavy-tailed weight distribution dominated by a few points, especially at the late denoising stage (visualized in \figref{fig:motivation}). 
To avoid such sharp weights and the resulting numerical instability, PCA adopts a biased weighted streaming softmax with batch-level averaging to manually flatten the weights. 
However, we find that it introduces a systematic smoothing bias that weakens the analytical denoiser and reduces fidelity.
As qualitatively evidenced in \figref{fig:wss_ss} (b) and the fourth row of \figref{fig:main_vis_results}, PCA struggles to recover high-frequency details and tends to produce blurry outputs even with sufficient denoising steps. 

Our approach is fundamentally different and more elegant: instead of correcting an ill-conditioned global weighting with a biased softmax, we first fix the support. Using \algopt, we dynamically filter a {\em Golden Subset}, a compact set of samples that are useful, information-rich, and reliable for the current noisy input. Once the support is purified in this way, we no longer need manual weight-flattening tricks: we can directly apply a simple and unbiased streaming softmax \cite{dao2022flashattention} on the {\em Golden Subset}. This naturally regulates the weight distribution, decoupling stability from bias, and yields sharper details and better semantic coherence without the over-smoothing effects of weighted softmax or the instability risks of naive full-corpus weighting.
As qualitatively demonstrated in \figref{fig:wss_ss}(c) and the fifth row of \figref{fig:main_vis_results}, the samples generated via \algopt closely align with those produced by neural denoisers.
A further ablation study and analysis are provided in \secref{sec:ablation}.

\vspace{-5pt}
\subsection{Why Dynamic Retrieval? A Sensitivity Analysis} \label{sec:method_sensitive}
To determine the optimal allocation of computational budgets, we investigate the mechanics of the estimator across the diffusion steps.
Combining the analysis of weights evolution in \figref{fig:weight_dist} and the denoiser's sensitivity to subset scaling in \figref{fig:mse_full_subset}, we characterize two distinct regimes governed by the Signal-to-Noise Ratio.

\noindent {\bf Early Stage: Global Manifold Approximation.}
In the early stages of diffusion, the signal is drowned out by noise ($\sigma_t^2 \gg 1$).
\figref{fig:weight_dist} shows a diffuse posterior where no single sample dominates.
Crucially, our sensitivity analysis in \figref{fig:mse_full_subset} reveals that performance drops significantly with small subsets ($N=10, 100$) but recovers with a larger random subset ($N=1000$).
Qualitatively, as shown in \figref{fig:vis_full_subset}, restricting the support to a small subset (e.g., $N=10$) during this phase yields reconstructions that manifest as blurred superpositions of the training samples, failing to capture the global data manifold.
Conversely, utilizing a larger subset produces outputs indistinguishable from those derived using the full dataset.
These observations imply that in this regime, the estimator functions as a Monte Carlo Integrator: it relies on the \textit{Law of Large Numbers} to approximate the global expectation $\mathbb{E}[\xx_0]$.
Therefore, the estimator is robust to \textit{retrieval imprecision} (random selection works if $N$ is large) but sensitive to \textit{sample sparsity}.
Thus, we need a \textit{broad support} ($k \uparrow$) to cover the manifold in this regime, but can rely on \textit{coarse retrieval}.

\noindent {\bf Late Stage: Local Neighbor Selection.}
As the noise vanishes ($\sigma_t^2 \to 0$), \figref{fig:weight_dist} demonstrates a sharp entropy collapse, where probability mass concentrates on a tiny neighborhood.
Here, the estimator shifts from \textit{integration} to \textit{selection}.
The Weight Gap between the true neighbor and distant samples explodes, meaning omitting the Top-1 neighbor introduces catastrophic bias.
The estimator becomes robust to \textit{sample sparsity} (a few samples suffice) but extremely sensitive to \textit{retrieval precision}.
Thus, we need \textit{high-precision retrieval} ($m \uparrow$) but can enforce \textit{aggressive sparsity} ($k \downarrow$).
Driven by this ``Integration-to-Selection'' transition, \algopt employs a time-aware mechanism that dynamically allocates the computational budget.
We propose a \textit{Counter-Monotonic Schedule} for retrieval scope ($m_t$) and aggregation budget ($k_t$).

\vspace{-5pt}
\subsection{Theoretical-grounded Dynamic Selection} \label{sec:dynamic_method}
\noindent{\bf Adaptive Coarse Screening.}
We first filter the full dataset $\mathcal{D}$ to a candidate set $\mathcal{C}_t$ of size $m_t$ using a computationally efficient proxy metric.
Specifically, we employ a spatially downsampled $\ell_2$ distance: $d^{\mathrm{proxy}}(\xx_t, \xx_i) \triangleq \|\operatorname{Down}_{s}(\xx_t) - \operatorname{Down}_{s}(\xx_i)\|_2$, where $s=1/4$.
This design leverages the \textit{hierarchical consistency} of natural images \cite{wang2004image}: distinct local similarity typically correlates with reasonable proximity in the low-frequency structure captured by our proxy.
Furthermore, to guarantee recall in the low-noise regime (where exact matches are critical), the candidate pool size $m_t$ must expand as noise decreases.
We model $m_t$ via a monotonically \textit{increasing} schedule with respect to signal strength:
\begin{equation}
    m_t = \lfloor m_{\min} + (m_{\max} - m_{\min}) \cdot (1 - g(\sigma_t)) \rfloor,
    \label{eq:m_schedule}
\end{equation}
where $g(\sigma_t) \in [0,1]$ is the normalized noise level.
This ensures $m_t \to m_{\max}$ as $t \to 0$, providing a ``safety margin'' to recall true neighbors when precision is paramount.

\noindent{\bf Precision Golden Set Selection .}
We then compute exact distances strictly within $\mathcal{C}_t$ to obtain the final golden support $S_t$ of size $k_t$:
\begin{equation}
    S_t = arg\,topk_{i \in \mathcal{C}_t, |S|=k_t} \left\{ -\|\xx_t - \xx_i\|_2 \right\}.
\end{equation}
Aligning with the posterior concentration (\figref{fig:motivation}), the aggregation budget $k_t$ follows an opposing schedule:
\begin{equation}
    k_t = \lfloor k_{\min} + (k_{\max} - k_{\min}) \cdot g(\sigma_t) \rfloor.
    \label{eq:k_schedule}
\end{equation}
As $t \to 0$, $k_t \to k_{\min}$, exploiting the sparsity of the posterior.
This symmetric trade-off minimizes complexity, achieving significant acceleration.
Our analysis in \secref{sec:hyper_analysis} demonstrates that the optimal values for these hyperparameters are empirically consistent across multiple datasets.

\begin{table}[!t]
\centering
\caption{\textbf{Algorithmic Complexity Comparison.}
$D$: flattened image dimension, $N$: dataset size, $p_t$: patch size at step $t$.}
\label{tab:complexity_results}
\resizebox{0.48\textwidth}{!}{%
    \begin{tabular}{lccccc}
        \toprule
        Method & Optimal & Wiener Filter & Kamb & PCA & \textbf{\algopt (Ours)} \\
        \midrule
        Complexity & $\mathcal{O}(ND)$ & $\mathcal{O}(D^2)$ & $\mathcal{O}(N p_t D^2)$ & $\mathcal{O}(N p_t D)$ & $\mathbf{\mathcal{O}(N d + m_t p_t D)}$ \\
        \bottomrule
    \end{tabular}}
    \vspace{-10pt}
\end{table}

\begin{figure*}[!t]
    \centering
    \includegraphics[width=\linewidth]{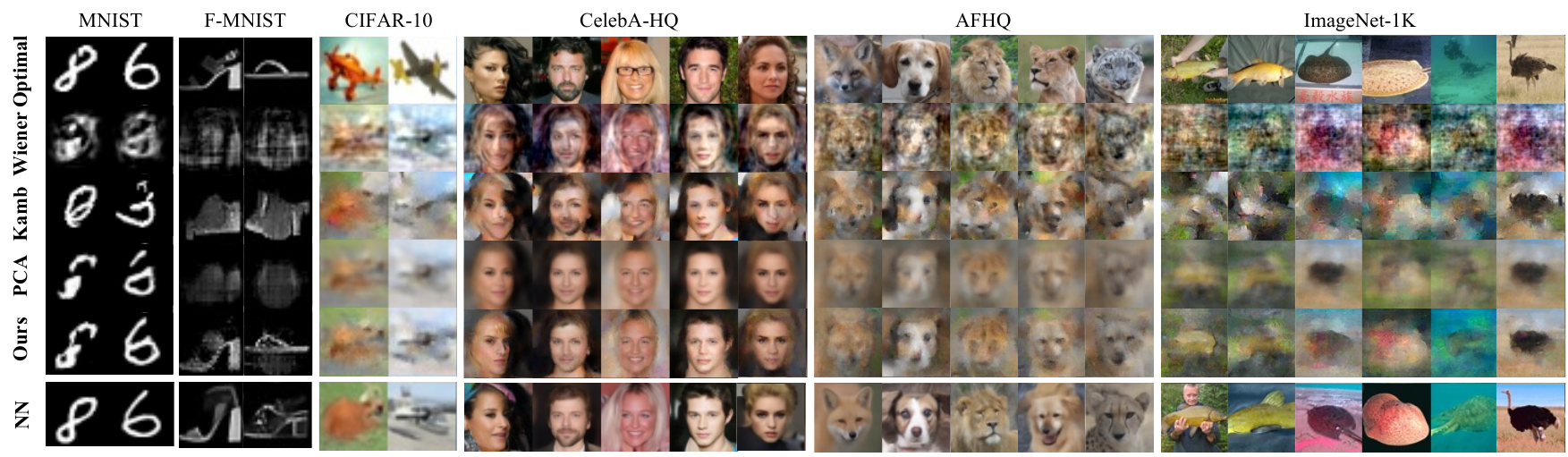}
    \caption{\textbf{Qualitative Comparison.}
    We compare our \algopt (5th row) against four baselines: Optimal (1st row), Wiener filter (2nd row), Kamb \cite{kamb2024analytic} (3rd row), and the PCA model (4th row).
    All images are generated from the same initial noise using 10 steps of DDIM \cite{song2020denoising}.
    The last row displays reference samples generated by a trained U-Net \cite{ho2020denoising}.}
    \label{fig:main_vis_results}
    \vspace{-5pt}
\end{figure*}

\vspace{-5pt}
\subsection{Truncation Error Bound and Complexity Analysis} \label{sec:complexity}
We rigorously bound the approximation error induced by truncating the support to a subset $S_t$.
Let $\hat{\ff}_{\cD}(\mathbf{x}_t)$ and $\hat{\ff}_{S_t}(\mathbf{x}_t)$ denote the exact and truncated posterior means, respectively.
We derive the following upper bound (see \appref{app:theory} for the detailed derivation):

\begin{theorem}[Posterior Truncation Error Bound] \label{thm:truncation_error}
Assume logits $\ell_i$ are sorted such that $\ell_{(1)} \ge \dots \ge \ell_{(N)}$.
If the estimator aggregates only the top-$k$ samples, the error is bounded by:
\begin{equation}
    \|\hat{\ff}_{\cD}(\xx_t)-\hat{\ff}_{S_t}(\xx_t)\|_2 \le 2R(N-k) \cdot \exp\left(-\Delta_k\right),
\end{equation}
where $R = \max_{\mathbf{x} \in \mathcal{D}} \|\mathbf{x}\|_2$ is the data radius, and $\Delta_k \triangleq \ell_{(1)} - \ell_{(k+1)}$ represents the \textit{Logit Gap}.
\end{theorem}

The bound reveals why our dynamic $k_t$ schedule is theoretically sound:
In the high-noise regime ($t \to T$), the noise term in the denominator of $\ell_i$ is large, suppressing differences between samples.
Consequently, the Logit Gap $\Delta_k \to 0$, and the exponential term $\exp(-\Delta_k) \to 1$.
The error bound becomes linear in $(N-k)$. To control the error, we must maximize $k$ (i.e., $k \to k_{\max}$), as prescribed by Eq. \eqref{eq:k_schedule}.
Conversely, in the low-noise regime ($t \to 0$), the Logit Gap widens significantly ($\Delta_k \gg 1$).
The error term $\exp(-\Delta_k)$ decays exponentially, rendering the tail contribution negligible.
This allows for aggressive truncation strategy ($k \to k_{\min}$), ensuring that computational efficiency is gained without performance loss.

We summarize the algorithmic complexity in \tabref{tab:complexity_results}.
Standard methods such as Kamb~\cite{kamb2024analytic} and PCA~\cite{Lukoianov2025Locality} typically scale linearly with the dataset size $N$ for every pixel or patch (e.g., $\mathcal{O}(N p_t D)$), rendering them computationally prohibitive for large-scale data.
By introducing a low-dimensional proxy space $\mathbb{R}^d$ (where $d \ll D$) and a dynamic search scope $m_t \ll N$, \algopt reduces the complexity.
Crucially, \algopt functions as a plug-and-play module and can be seamlessly integrated into existing baselines (detailed in \secref{sec:orthogonal}).
For instance, when applied to the PCA\cite{Lukoianov2025Locality}, it effectively reduces the complexity to $\mathcal{O}(N d + m_t p_t D)$.

\begin{table*}[t]
\centering
\caption{\textbf{Quantitative Comparison of Analytical Denoisers.} 
Here, \emph{Time} denotes the sampling time cost (s) per step, and \emph{Memory} denotes the peak memory (GB) usage during sampling.
All metrics are averaged over 128 samples.
The best and second-best results are highlighted in \textbf{bold} and \underline{underlined}, respectively.
Efficacy gain (\%) denotes the improvement relative to the second-best performance.
}
\vspace{-5pt}

\label{tab:main_results_1}
\resizebox{\textwidth}{!}{%
    \begin{tabular}{lcccccccccccc}
    \toprule
    Dataset 
    & \multicolumn{4}{c}{\textbf{CIFAR-10}} 
    & \multicolumn{4}{c}{\textbf{CelebA-HQ}}
    & \multicolumn{4}{c}{\textbf{AFHQ}}\\
    \cmidrule(lr){2-5} \cmidrule(lr){6-9} \cmidrule(lr){10-13}
    
    Metric
    & \multicolumn{2}{c}{Efficacy} & \multicolumn{2}{c}{Efficiency}
    & \multicolumn{2}{c}{Efficacy} & \multicolumn{2}{c}{Efficiency}
    & \multicolumn{2}{c}{Efficacy} & \multicolumn{2}{c}{Efficiency} \\
    \cmidrule(lr){2-3} \cmidrule(lr){4-5}
    \cmidrule(lr){6-7} \cmidrule(lr){8-9}
    \cmidrule(lr){10-11} \cmidrule(lr){12-13}
    
    Method
    & MSE ($\downarrow$) & $r^2$ ($\uparrow$) & Time & Memory
    & MSE ($\downarrow$) & $r^2$ ($\uparrow$)  & Time & Memory
    & MSE ($\downarrow$) & $r^2$ ($\uparrow$)  & Time & Memory \\
    \midrule
    
    Optimal \cite{de2022convergence} & 0.030 & -0.401 & 2.353 & 0.020  & 0.021 & 0.010 & 1.244 & 0.055 & 0.039 & -0.713 & 0.642 & 0.055  \\
    Wiener \cite{wiener1949extrapolation}  & 0.008 & 0.553  & 0.011 & 0.290 & 0.012 & 0.781 & 0.153 & 4.509  & 0.010 & 0.687  & 0.156 & 4.509  \\
    Kamb \cite{kamb2024analytic}    & 0.011 & 0.411  & 5.980 & 1.889 & 0.011 & 0.720 & 37.158 & 7.547 & 0.034 & 0.576  & 15.094 & 7.5471 \\\midrule
     PCA~\cite{Lukoianov2025Locality}     & \underline{0.008} & \underline{0.670}  & 2.802 & 0.835 & \underline{0.009} & \underline{0.802} & 6.040 & 5.460 & \underline{0.008} & 0.703 & 24.896 & 4.743 \\
    \hl{\algopt (Ours)}
    & \hl{\textbf{0.007}}
    & \hl{\textbf{0.683}}
    & \hl{0.087}
    & \hl{0.882}
    
    & \hl{\textbf{0.008}}
    & \hl{\textbf{0.836}}
    & \hl{0.349} 
    & \hl{{5.545}} 
    
    & \hl{\textbf{0.007}}
    & \hl{\textbf{0.731}}
    & \hl{0.351}
    & \hl{{4.783}}\\
    \textit{vs. PCA} & \textcolor{darkgreen}{$\uparrow$ 12.5\%} & \textcolor{darkgreen}{$\uparrow$ 1.9\%} & \textcolor{darkgreen}{\boldmath$\times \textbf{28.1}$} & - & \textcolor{darkgreen}{$\uparrow$ 11.1\%} & \textcolor{darkgreen}{$\uparrow$ 4.2\%}  & \textcolor{darkgreen}{\boldmath$\times$ \textbf{17.4}} & - & \textcolor{darkgreen}{$\uparrow$ 12.5\%} & \textcolor{darkgreen}{$\uparrow$ 3.9\%} & \textcolor{darkgreen}{\boldmath$\times$ \textbf{71.0}} & - \\
\bottomrule
    \end{tabular}}
    \vspace{-10pt}
\end{table*}

\vspace{-5pt}
\section{Experiments}
\vspace{-5pt}
Our evaluation is guided by three core questions: (1) \textit{Efficacy}: Does the dynamic ``Golden Subset'' approximation converge to the exact analytical score derived from the full dataset?
(2) \textit{Efficiency}: Can \algopt significantly reduce the computational complexity?
(3) \textit{Scalability}: Can \algopt effectively scale to large-scale benchmarks?

\vspace{-5pt}
\subsection{Experimental Setup}
\noindent \textbf{Datasets.}
We employ a comprehensive evaluation protocol spanning diverse resolutions and complexities.
Benchmarks include standard grayscale datasets (MNIST \cite{deng2012mnist}, FashionMNIST \cite{xiao2017fashion}; $28 \times 28$), natural images (CIFAR-10 \cite{krizhevsky2009learning}; $32 \times 32$), and high-resolution structured domains (CelebA-HQ \cite{karras2017progressive}, AFHQv2 \cite{choi2020stargan}; $64 \times 64$).
Crucially, we extend analytical denoising evaluation to the large-scale ImageNet-1K dataset \cite{deng2009imagenet} ($64 \times 64$) for the first time, reporting results of both unconditional and class-conditional generation.

\noindent{\bf Baselines.}
We compare \algopt against a hierarchy of analytical and neural denoisers:
\begin{itemize}[leftmargin=*, itemsep=0pt, topsep=0pt]
\item \textit{Analytical Denoisers:} We benchmark against the classical Wiener filter \cite{wiener1949extrapolation}, the Optimal Denoiser \cite{de2022convergence}, the patch-based method by Kamb et al.\footnote{We refer to this method as ``Kamb'' for brevity, as the original paper does not propose a specific method name.} \cite{kamb2024analytic}, and the state-of-the-art PCA denoiser \cite{Lukoianov2025Locality}.
\item \textit{Neural Denoisers:} To assess the efficacy of analytical denoisers, following \cite{kamb2024analytic,Lukoianov2025Locality}, we utilize the DDPM U-Net \cite{ho2020denoising} (with self-attention removed for fair architectural comparison) as the ground truth oracle.
Furthermore, we extend the evaluation protocol by additionally employing the strong EDM \cite{karras2022elucidating}.
\end{itemize}

\noindent{\bf Evaluation Metrics.}
We assess performance along two axes:
\textit{(1) Efficacy:} Following previous works \cite{kamb2024analytic,Lukoianov2025Locality}, we report Mean Squared Error (MSE) and the coefficient of determination ($r^2$) to quantify alignment between the analytical score estimate and the neural oracle.
Higher $r^2$ ($\uparrow$) and lower MSE ($\downarrow$) indicate better efficacy.
\textit{(2) Efficiency:} We measure time (s) per denoising step and peak memory usage (GB).

\noindent{\bf Implementation Details.}
\algopt operates as a plug-and-play module that can be seamlessly integrated into existing analytical diffusion denoisers.
To demonstrate this versatility while ensuring a rigorous comparison with the state-of-the-art, we primarily report results deploying \algopt atop the PCA denoiser \cite{Lukoianov2025Locality}.
Additional results demonstrating its universality to other baselines are detailed in \secref{sec:orthogonal}.
Regarding the subset selection, the coarse set is dynamically expanded from $m_{min}$ to $m_{max}$, and the ``Golden Subset'' is reduced from $k_{max}$ to $k_{min}$ as the denoising process.
In practice, for $m_{min}$ and $k_{max}$, a large random subset (e.g., 5,000 samples for CIFAR-10) achieves performance parity with the full dataset (\figref{fig:mse_full_subset}), thus we set $m_{min}=k_{max}=\nicefrac{N}{10}$ where $N$ denotes the size of training set.
We set $m_{max}=\nicefrac{N}{4}$ and $k_{\min}=\nicefrac{N}{20}$ for all datasets.
A detailed sensitivity analysis (\secref{sec:hyper_analysis}) confirms that these hyperparameters are empirically consistent across multiple datasets.
Following \cite{Lukoianov2025Locality}, the number of diffusion steps is set to $10$ by default.

\vspace{-5pt}
\subsection{Efficacy and Efficiency Comparison}
We evaluate the performance of \algopt against established analytical baselines across multiple datasets, including the small-scale datasets and the large-scale dataset ImageNet-1K \cite{deng2009imagenet}.

\noindent{\bf Results on Small-scale Datasets.}
Quantitative metrics and qualitative comparisons are provided in \tabref{tab:main_results_1} and \figref{fig:main_vis_results}, respectively, with additional results for MNIST and Fashion-MNIST detailed in \appref{app_sec:exp}.

\textbf{\textit{Efficacy.}}
As evidenced in \tabref{tab:main_results_1}, \algopt consistently surpasses the state-of-the-art PCA method \cite{Lukoianov2025Locality} across all MSE and $r^2$ metrics.
This suggests that our dynamic subset selection effectively filters out irrelevant samples, yielding more reliable score estimates.
Qualitatively, \figref{fig:main_vis_results} corroborates these quantitative gains: compared to competing baselines, \algopt exhibits superior similarity to the ground-truth neural diffusion outputs (see last row).
Notably, the optimal denoiser (1st row) produces visually clear outputs.
However, this stems from merely ``memorizing'' \cite{kamb2024analytic} the training samples and fails to generalize, leading to significantly inferior MSE and $r^2$ metrics compared to our method.

\textbf{\textit{Efficiency.}}
Our \algopt significantly accelerates inference, achieving speedup factors of \textbf{17 $\times$} and \textbf{71 $\times$} on the CelebA-HQ and AFHQ datasets, respectively, compared to the SOTA PCA method \cite{Lukoianov2025Locality}, while maintaining a comparable memory cost.
In contrast, existing baselines struggle to balance inference speed and quality.
While the Wiener filter is computationally efficient because its complexity depends solely on image dimensions rather than dataset size (see \tabref{tab:complexity_results}), it suffers from markedly inferior performance.
On the other hand, Kamb et al. \cite{kamb2024analytic} incur prohibitive computational and memory costs.
By assuming translation equivariance, this method necessitates comparing the patch surrounding each pixel (of size $p_t$) against every patch in the training dataset, leading to substantial memory overhead.
Furthermore, it relies on a heuristic dependency that requires estimating the effective receptive field of a pre-trained U-Net to determine the patch size $p_t$ for each diffusion timestep, which introduces considerable additional computational burden.

\begin{table}[!t]
\centering
\caption{\textbf{Quantitative Comparison on ImageNet-1K.} We evaluate performance in both unconditional and conditional settings. All metrics are averaged over 128 samples, with the best results highlighted in \textbf{bold}. ``Total Steps'' denotes the total number of denoising steps, and we report results for 10 and 100 steps.}

\label{tab:pca_results_large}
\resizebox{.48\textwidth}{!}{%
    \begin{tabular}{llcccccc}
    \toprule
     \multirow{2}{*}{\makecell[c]{Total\\Steps}} 
    & \multirow{2}{*}{Method}
    & \multicolumn{3}{c}{\textbf{Unconditional}}
    & \multicolumn{3}{c}{\textbf{Conditional}} \\
    \cmidrule(lr){3-5} \cmidrule(lr){6-8}
    & & MSE ($\downarrow$) & $r^2$ ($\uparrow$) & Time
    & MSE ($\downarrow$) & $r^2$ ($\uparrow$) & Time\\
    \midrule
    \multirow{3}{*}{10}
    & PCA & 0.045  & 0.412  & \underline{110.798} &  0.033 & 0.467 & \underline{0.451}\\
    & PCA (Unbaised) & \underline{0.042}  & \underline{0.435}  & 110.986 &  \underline{0.032} & \underline{0.473} & 0.453\\
    & \hl{\algopt} & \hl{\textbf{0.039}} & \hl{\textbf{0.458}} & \hl{\textbf{2.607}} & \hl{\textbf{0.031}} & \hl{\textbf{0.490}} & \hl{\textbf{0.100}} \\
    
    \midrule
    \multirow{3}{*}{100}
    & PCA & \underline{0.032} & \underline{0.436}  & \underline{110.798} & \underline{0.025} & \underline{0.521} & \underline{0.451}\\
    & PCA (Unbaised) & 0.036  &0.425 & 110.986 &  0.035 & 0.411 & 0.453\\
    & \hl{\algopt} & \hl{\textbf{0.027}} & \hl{\textbf{0.509}} & \hl{\textbf{2.607}} & \hl{\textbf{0.022}} & \hl{\textbf{0.578}}& \hl{\textbf{0.100}} \\
\bottomrule
\end{tabular}}
\vspace{-5pt}
\end{table}

\noindent{\bf Results on Large-scale ImageNet-1K.}
We evaluate \algopt on large-scale ImageNet-1K under both unconditional and conditional settings, employing EDM~\cite{karras2022elucidating} as the neural denoiser.
Building upon the quantitative superiority reported in \tabref{tab:main_results_1}, we benchmark our method against the SOTA PCA~\cite{Lukoianov2025Locality} and its unbiased variant, \textit{PCA (Unbiased)}, which uses streaming softmax~\cite{dao2022flashattention} for weight estimation (detailed in \secref{sec:revisit_weight} and further analysis provided in \secref{sec:ablation}).

For \textit{Unconditional Generation}, as reported in \tabref{tab:pca_results_large}, \algopt consistently achieves state-of-the-art performance across varying sampling budgets (total denoising steps $T=10$ and $T=100$).
Moreover, in terms of efficiency, our method achieves an approximate $42\times$ acceleration in inference compared to baselines.
Note that we report the inference time (s) per denoising step; thus, this per-step cost remains constant regardless of the total budget $T$.

\begin{figure}[!t]
    \centering
    \includegraphics[width=\linewidth]{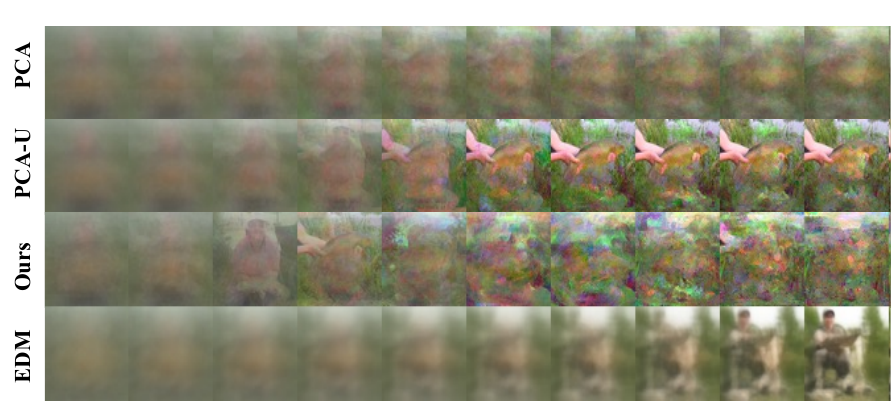}
    \caption{\textbf{Qualitative Comparison of conditional Denoising on ImageNet-1K.} 
    We visualize samples generated by our method compared to two PCA-based baselines: Original PCA \cite{Lukoianov2025Locality}
    and Unbiased PCA (PCA-U) for the class ``Tench''.}
    \label{fig:vis_imagenet_1k}
\end{figure}

For \textit{Conditional Generation,} 
\tabref{tab:pca_results_large} provides the detailed quantitative results, where we report the mean performance and inference time averaged across all classes.
Obviously, our method achieves superior performance combined with significant inference efficiency.
Furthermore, we visualize the intermediate denoised images for the class ``Tench'' in \figref{fig:vis_imagenet_1k}.
We reveal that both PCA-based baselines using the full dataset suffer from distinct failure modes rooted in their weight aggregation heuristics: (1) vanilla PCA uses the weight-averaging strategy, which leads to an \textit{over-smoothed} distribution and then causes the denoiser to suppress high-frequency details. 
This observation is consistent with the unconditional generation trends on other datasets in \figref{fig:wss_ss}.
(2) PCA (Unbiased) introduces a pathological bias toward specific training exemplars.
This manifests as explicit data memorization rather than genuine manifold learning: as observed in the second row, the outputs appear as disjointed "patch-collages," often marred by anatomical incoherence (e.g., the unnatural intrusion of human hands from the training set).
Crucially, increasing the denoising steps $T$ exacerbates this patch-pasting behavior, causing visual artifacts to stabilize rather than resolve.
The conclusion is also verified by the results in \tabref{tab:pca_results_large}, where both MSE and $r^2$ metrics decrease from $T=10$ to $T=100$ for conditional generation.

\begin{table}[!t]
\centering
\setlength{\tabcolsep}{10pt}
\caption{\textbf{Validation on Diverse Neural Denoisers.}
All metrics are averaged over 128 samples.}
\label{tab:more_nn}
\resizebox{.48\textwidth}{!}{%
\begin{tabular}{llcccc}
\toprule

& 
& \multicolumn{2}{c}{\textbf{CIFAR-10}}
& \multicolumn{2}{c}{\textbf{AFHQ}} \\
\cmidrule(lr){3-4} \cmidrule(lr){5-6}
& & MSE ($\downarrow$) & $r^2$ ($\uparrow$) 
& MSE ($\downarrow$) & $r^2$ ($\uparrow$) \\ \midrule

\multirow{5}{*}{EDM-VP}
& Optimal & 0.060 & -0.238 & 0.053 &  0.028 \\
& Wiener  & 0.026 & 0.472 & 0.033 & 0.448\\
& Kamb &  0.031 & 0.350 & 0.034 &0.461 \\
& PCA & \underline{0.023} & \underline{0.594} & \underline{0.029} & \underline{0.524}\\
& \hl{\algopt} & \hl{\textbf{0.0195}} & \hl{\textbf{0.654}} & \hl{\textbf{0.024}} & \hl{\textbf{0.588}} \\
 \midrule 
 
\multirow{5}{*}{EDM-VE}
& Optimal & 0.057 & -0.226 &  0.051 & 0.034\\
& Wiener & 0.026 & 0.461 & 0.031 & 0.512 \\
& Kamb & 0.029 & 0.392 & 0.035 & 0.459 \\
& PCA & \underline{0.022} & \underline{0.604} & \underline{0.029} & \underline{0.535}\\
& \hl{\algopt} & \hl{\textbf{0.018}} & \hl{\textbf{0.666}} & \hl{\textbf{0.023}} & \hl{\textbf{0.595}} \\

\bottomrule
\end{tabular}}
\vspace{-10pt}
\end{table}

In contrast, our \algopt is fundamentally different and more elegant: we dynamically filter a {\em Golden Subset}, a compact set of samples that are useful, information-rich, and reliable for the current noisy input. 
Restricted to this optimized support set, this weight distribution can strike a critical balance: it circumvents the catastrophic smoothing of vanilla PCA while mitigating the memorization traps in the unbiased variant for scanning the full dataset.
Consequently, as shown in \figref{fig:main_vis_results}, our method yields images characterized by sharper details and better semantic coherence without the over-smoothing effects of weighted streaming softmax or the instability risks of naive full-corpus weighting.
Results reported in \tabref{tab:pca_results_large} also demonstrate its superiority.

\noindent{\bf Validation on Diverse Network Denoisers.}
We assess the generality of our method \algopt by comparing it against the outputs of the state-of-the-art network-based denoisers, including EDM-VP and EDM-VE~\cite{karras2022elucidating}.
As illustrated in~\tabref{tab:more_nn}, our method \algopt consistently outperforms the PCA baseline in matching the generation quality of these networks.
This confirms that \algopt provides a more robust and high-quality analytical model that better aligns with the complex manifolds learned by deep neural networks.

\begin{table}[!t]
\centering
\caption{\textbf{Orthogonality to Existing Analytical Denoisers.}
All metrics are averaged over 128 samples.}
\label{tab:orthogonal}
\resizebox{.48\textwidth}{!}{%
    \begin{tabular}{l ccc ccc}
    \toprule
    & \multicolumn{3}{c}{\textbf{Celeba-HQ}} & \multicolumn{3}{c}{\textbf{AFHQ}} \\
    \cmidrule(lr){2-4} \cmidrule(lr){5-7}
    \textbf{Method} & MSE ($\downarrow$) & $r^2$ ($\uparrow$) & Time & MSE ($\downarrow$) & $r^2$ ($\uparrow$) & Time \\ 
    \midrule
    Optimal    & 0.021 & 0.010 & 1.244 & 0.039 & -0.713 & 0.642 \\
    \rowcolor[HTML]{F2F2F2}
    + \algopt  & \textbf{0.014} & \textbf{0.134} & \textbf{0.338}  & \textbf{0.035} & \textbf{-0.270}  & \textbf{0.153}  \\
    \midrule
    Kamb       & 0.011 & 0.720 & 37.158 & 0.034 & 0.576  & 15.094 \\
    \rowcolor[HTML]{F2F2F2}
    + \algopt  & \textbf{0.009} & \textbf{0.729} & \textbf{1.336} & \textbf{0.031} & \textbf{0.586} & \textbf{1.434} \\
    \bottomrule
    \end{tabular}
    \vspace{-10pt}
}
\end{table}

\noindent {\bf Orthogonality to Existing Analytical Denoisers.}\label{sec:orthogonal}
We further investigate the compatibility of \algopt by incorporating it into other representative analytical baselines, including the Optimal Denoiser~\cite{de2022convergence} and Kamb~\cite{kamb2024analytic}.
Note that the Wiener filter~\cite{wiener1949extrapolation} is excluded from this analysis, as it relies solely on pre-computed statistics and does not require accessing the explicit training corpus during sampling.
Results in \tabref{tab:orthogonal} demonstrate that equipping these methods with \algopt not only yields superior performance but also achieves significant acceleration.
This confirms that our approach acts as a versatile plug-and-play enhancement orthogonal to the current analytical solvers.

\subsection{Ablation Study}\label{sec:ablation}
\noindent {\bf Impact of Biased Weight Estimation.}
To quantify the impact of the weight estimation mechanism discussed in \secref{sec:revisit_weight}, we conduct an ablation study focusing on its role during the denoising process. Specifically, we evaluate \algopt under two configurations: (1) employing the biased Weighted Streaming Softmax (WSS), and (2) utilizing the Unbiased Streaming Softmax~\cite{dao2022flashattention} (SS).
The results, reported in \tabref{tab:ablation}, demonstrate that biased weighting leads to performance degradation, whereas the unbiased formulation consistently yields superior denoising results.
This is attributed to the efficacy of \algopt: by explicitly filtering a ``Golden Subset'' that is highly reliable and information-rich for the current noisy input, the unbiased estimator can effectively reconstruct the posterior.

\begin{figure}[!t]
    \centering
     \begin{subfigure}[t]{0.238\textwidth}
        \centering
        \includegraphics[width=\linewidth]{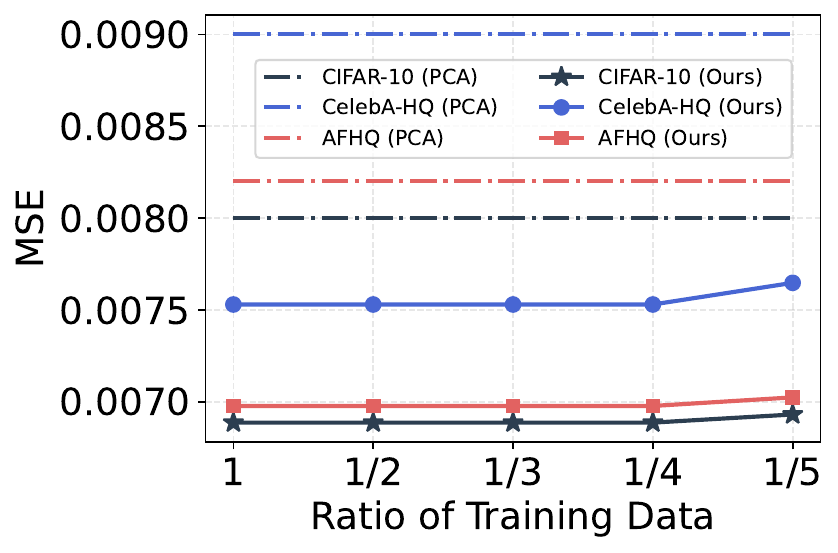}
        \caption{Coarse Set $m_{max}$}
        \vspace{-3pt}
        \label{fig:impact_m}
    \end{subfigure}
     \begin{subfigure}[t]{0.238\textwidth}
        \centering
        \includegraphics[width=\linewidth]{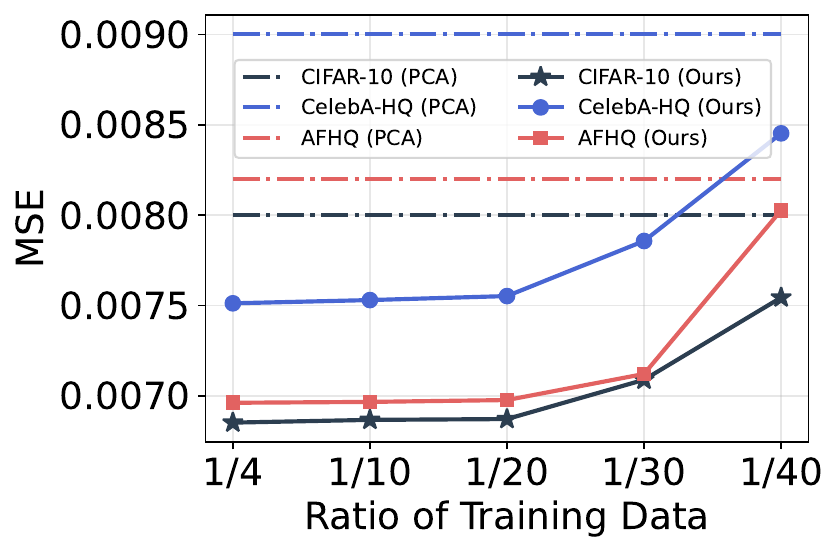}
        \caption{Golden Subset $k_{min}$}
        \vspace{-3pt}
        \label{fig:impact_k}
    \end{subfigure}
    \caption{\textbf{Sensitivity Analysis} of the coarse candidate set size and golden subset size across multiple datasets.
    Dashed lines mean the baseline PCA \cite{Lukoianov2025Locality}.}
    \label{fig:hyper_analysis}
\end{figure}

\begin{table}[!t]
\centering
\setlength{\tabcolsep}{12pt}
\caption{Biased (WSS) vs. unbiased weight (SS) estimation.
All metrics are averaged over 128 samples.}
\label{tab:ablation}
\resizebox{.47\textwidth}{!}{%
\begin{tabular}{llcccc}
\toprule

& 
& \multicolumn{2}{c}{\textbf{Celaba-HQ}}
& \multicolumn{2}{c}{\textbf{AFHQ}} \\
\cmidrule(lr){3-4} \cmidrule(lr){5-6}
& & MSE ($\downarrow$) & $r^2$ ($\uparrow$) 
& MSE ($\downarrow$) & $r^2$ ($\uparrow$)  \\ \midrule

\multirow{2}{*}{\algopt}
& + WSS & 0.009 & 0.818 & 0.008  & 0.715 \\
& +SS  & \hl{\textbf{0.008}} & \hl{\textbf{0.836}} & \hl{\textbf{0.007}} & \hl{\textbf{0.731}}\\ 
\bottomrule
\end{tabular}}
\vspace{-10pt}
\end{table}

\noindent {\bf Hyperparameter Analysis.}\label{sec:hyper_analysis}
We perform a sensitivity analysis on two critical hyperparameters: the maximum coarse subset size ($m_{\max}$) and the minimum golden subset size ($k_{\min}$).
For \textit{Coarse Subset Size ($m_{\max}$)}: This parameter controls the late-stage denoising by determining the candidate pool size required to identify precise nearest neighbors.
We evaluate $m_{\max} \in \{1, \nicefrac{1}{2}, \nicefrac{1}{3}, \nicefrac{1}{4}, \nicefrac{1}{5}\}$ relative to the full training dataset size $N$.
As illustrated in \figref{fig:impact_m}, \algopt exhibits remarkable consistency across multiple datasets.
To balance computational overhead and approximation accuracy, we empirically set $m_{\max} = \nicefrac{4}{N}$ as the default.
Notably, when $m_{\max}$ is smaller, performance degrades significantly, as the subset may fail to encompass the optimal neighborhood for the subsequent golden subset selection.
For \textit{Golden Subset Size ($k_{\min}$):} We further investigate the impact of the fine-subset selection bounds.
We evaluate $k_{\min}$ across the set $\{\nicefrac{1}{4}, \nicefrac{1}{10}, \nicefrac{1}{20}, \nicefrac{1}{30}, \nicefrac{1}{40}\}$ relative to training data size $N$.
We empirically set $k_{\min} = \nicefrac{20}{N}$.
The results in \figref{fig:impact_k} demonstrate consistent performance across multiple datasets.
Similar to the behavior observed with the coarse subset, a performance drop occurs at the extreme lower bound, where the subset becomes too sparse to provide sufficient local guidance for the diffusion process.

\section{Conclusion}
We resolved the scalability bottleneck of analytical diffusion by transforming a mathematically transparent yet prohibitive framework into a practical engine for large-scale generative modeling.
Through the lens of \textit{Posterior Progressive Concentration}, we proved that global full training set scanning is often redundant and even harmful.
Our \algopt framework dynamically filters a {\em Golden Subset} via a theoretical-grounded dynamical selection, effectively decouples inference complexity from dataset scale.
This approach achieves a $71\times$ speedup on AFHQ and represents the first successful scaling of analytical diffusion to ImageNet-1K.
Furthermore, we establish a scalable training-free paradigm that bridges the gap between mathematical transparency and industrial-scale generation.

\clearpage
\section*{Impact Statement}
This paper presents work whose goal is to advance the field of Machine Learning.
There are many potential societal consequences of our work, none which we feel must be specifically highlighted here.

\bibliography{resources/main_ref}
\bibliographystyle{configuration/icml2026}


\newpage
\appendix
\onecolumn

\section*{Appendix}

\section{Theoretical Analysis and Proofs} \label{app:theory}

In this section, we provide the rigorous derivation of the approximation error bound presented in \thmref{thm:truncation_error} and analyze the asymptotic behavior of our estimator in the high-noise and low-noise regimes.

\subsection{Proof of Theorem \ref{thm:truncation_error}}

\textbf{Problem Setup.}
Recall that the exact analytical denoiser $\hat{\ff}_{\cD}(\xx_t)$ is given by a softmax-weighted average over the entire dataset $\cD$:
\begin{equation}
    \hat{\ff}_{\cD}(\xx_t) = \sum_{i=1}^N w_i \xx_i, \quad w_i = \frac{\exp(\ell_i)}{Z}, \quad Z = \sum_{j=1}^N \exp(\ell_j),
\end{equation}
where $\ell_i = -\| \xx_t/\sqrt{\alpha_t} - \xx_i \|^2 / 2\sigma_t^2$ are the logits, and $Z$ is the partition function.

Let $S_t$ be the subset of indices corresponding to the top-$k$ largest logits. Without loss of generality, assume the indices are sorted such that $\ell_{(1)} \ge \ell_{(2)} \ge \dots \ge \ell_{(N)}$. Thus, $S_t = \{(1), \dots, (k)\}$.
The truncated estimator $\hat{\ff}_{S_t}(\xx_t)$ re-normalizes the weights over $S_t$:
\begin{equation}
    \hat{\ff}_{S_t}(\xx_t) = \sum_{i \in S_t} \tilde{w}_i \xx_i, \quad \tilde{w}_i = \frac{\exp(\ell_i)}{Z_S}, \quad Z_S = \sum_{j \in S_t} \exp(\ell_j),
\end{equation}
where $Z_S$ is the partial partition function.

\textbf{Step 1: Decomposition of the Error.}
The error vector is the difference between the exact and truncated estimates:
\begin{align}
    \mathbf{e} &= \hat{\ff}_{S_t}(\xx_t) - \hat{\ff}_{\cD}(\xx_t) \\
    &= \sum_{i \in S_t} \tilde{w}_i \xx_i - \left( \sum_{i \in S_t} w_i \xx_i + \sum_{i \notin S_t} w_i \xx_i \right) \\
    &= \sum_{i \in S_t} (\tilde{w}_i - w_i) \xx_i - \sum_{i \notin S_t} w_i \xx_i.
\end{align}

\textbf{Step 2: Analyzing the Weight Difference.}
Let $Z_{tail} = Z - Z_S = \sum_{j=k+1}^N \exp(\ell_{(j)})$ be the residual mass of the truncated samples.
For any index $i \in S_t$, the relationship between the exact weight $w_i$ and truncated weight $\tilde{w}_i$ is:
\begin{equation}
    w_i = \frac{e^{\ell_i}}{Z} = \frac{e^{\ell_i}}{Z_S + Z_{tail}} = \frac{e^{\ell_i}}{Z_S} \cdot \frac{Z_S}{Z_S + Z_{tail}} = \tilde{w}_i \left( 1 - \frac{Z_{tail}}{Z} \right).
\end{equation}
Thus, the difference is:
\begin{equation}
    \tilde{w}_i - w_i = \tilde{w}_i - \tilde{w}_i \left( 1 - \frac{Z_{tail}}{Z} \right) = \tilde{w}_i \frac{Z_{tail}}{Z}.
\end{equation}
Note that this difference is always positive, as renormalization increases the weights of retained samples.

\textbf{Step 3: Bounding the Norm.}
Applying the triangle inequality to the error vector $\mathbf{e}$:
\begin{align}
    \|\mathbf{e}\|_2 &\le \sum_{i \in S_t} |\tilde{w}_i - w_i| \|\xx_i\|_2 + \sum_{i \notin S_t} |w_i| \|\xx_i\|_2 \\
    &\le R \left( \sum_{i \in S_t} (\tilde{w}_i - w_i) + \sum_{i \notin S_t} w_i \right),
\end{align}
where $R = \max_i \|\xx_i\|_2$ is the bound on the data norm.
Substituting the expressions from Step 2:
\begin{itemize}
    \item First term sum: $\sum_{i \in S_t} (\tilde{w}_i - w_i) = \sum_{i \in S_t} \tilde{w}_i \frac{Z_{tail}}{Z} = \frac{Z_{tail}}{Z} \underbrace{\sum_{i \in S_t} \tilde{w}_i}_{1} = \frac{Z_{tail}}{Z}$.
    \item Second term sum: $\sum_{i \notin S_t} w_i = \frac{Z_{tail}}{Z}$ (by definition of $Z_{tail}$).
\end{itemize}
Thus, the total error is bounded by:
\begin{equation}
    \|\mathbf{e}\|_2 \le R \left( \frac{Z_{tail}}{Z} + \frac{Z_{tail}}{Z} \right) = 2R \frac{Z_{tail}}{Z}.
    \label{eq:error_ratio}
\end{equation}

\textbf{Step 4: Bounding the Ratio with Logit Gap.}
We now bound the ratio $Z_{tail}/Z$ using the sorted logits.
\begin{itemize}
    \item Lower bound for $Z$: Since $\ell_{(1)}$ is the maximum logit, $Z > \exp(\ell_{(1)})$.
    \item Upper bound for $Z_{tail}$: The tail consists of $N-k$ samples, each with logit $\ell_{(j)} \le \ell_{(k+1)}$ for $j > k$. Thus, $Z_{tail} = \sum_{j=k+1}^N e^{\ell_{(j)}} \le (N-k) e^{\ell_{(k+1)}}$.
\end{itemize}
Substituting these into Eq. \eqref{eq:error_ratio}:
\begin{equation}
    \frac{Z_{tail}}{Z} \le \frac{(N-k) e^{\ell_{(k+1)}}}{e^{\ell_{(1)}}} = (N-k) \exp\left( -(\ell_{(1)} - \ell_{(k+1)}) \right).
\end{equation}
Defining the Logit Gap as $\Delta_k \triangleq \ell_{(1)} - \ell_{(k+1)}$, we obtain the final bound:
\begin{equation}
    \|\hat{\ff}_{\cD}(\xx_t)-\hat{\ff}_{S_t}(\xx_t)\|_2 \le 2R(N-k) \exp(-\Delta_k).
\end{equation}
\qed


\subsection{Asymptotic Analysis of Regime Dynamics} \label{app:asymptotic}

Here, we analyze the behavior of the Logit Gap $\Delta_k$ as a function of the noise level $\sigma_t$, providing the mathematical justification for the "Integration-to-Selection" phase transition discussed in \secref{sec:method}.

Let the distance between the noisy query $\xx_t$ and a training sample $\xx_i$ be $d_i = \| \xx_t/\sqrt{\alpha_t} - \xx_i \|^2$. The logit is given by $\ell_i = -d_i / 2\sigma_t^2$.
The Logit Gap becomes:
\begin{equation}
    \Delta_k(\sigma_t) = \frac{d_{(k+1)} - d_{(1)}}{2\sigma_t^2}.
\end{equation}
Let $\delta_k = d_{(k+1)} - d_{(1)} > 0$ be the raw distance gap between the top-1 and the $(k+1)$-th neighbor.

\paragraph{Case 1: High-Noise Regime ($\sigma_t^2 \to \infty$).}
In the early stages of reverse diffusion, $\sigma_t^2$ is very large.
\begin{equation}
    \lim_{\sigma_t^2 \to \infty} \Delta_k(\sigma_t) = \lim_{\sigma_t^2 \to \infty} \frac{\delta_k}{2\sigma_t^2} = 0.
\end{equation}
Consequently, the exponential decay term $\exp(-\Delta_k) \to e^0 = 1$.
The error bound simplifies to:
\begin{equation}
    \text{Error} \lesssim 2R(N-k).
\end{equation}
\textbf{Implication:} The error is linearly proportional to the number of discarded samples $(N-k)$. To minimize error, we must minimize $(N-k)$, which implies maximizing $k$ (i.e., $k \to N$). This proves that \textit{dense aggregation} is mathematically necessary in the high-noise regime.

\paragraph{Case 2: Low-Noise Regime ($\sigma_t^2 \to 0$).}
As the diffusion process nears the data manifold, $\sigma_t^2 \to 0$. Assuming the query $\xx_t$ is not equidistant to $\xx_{(1)}$ and $\xx_{(k+1)}$ (which holds almost surely for continuous data), we have $\delta_k > 0$.
\begin{equation}
    \lim_{\sigma_t^2 \to 0} \Delta_k(\sigma_t) = \lim_{\sigma_t^2 \to 0} \frac{\delta_k}{2\sigma_t^2} = +\infty.
\end{equation}
Consequently, the exponential term vanishes rapidly:
\begin{equation}
    \lim_{\sigma_t^2 \to 0} \exp(-\Delta_k) = 0.
\end{equation}
Crucially, this decay is exponential in $1/\sigma_t^2$, which dominates the linear term $(N-k)$.
Even if we discard the vast majority of the dataset (i.e., $N-k \approx N$, $k$ is small), the total error remains negligible. This proves that \textit{sparse selection} is theoretically sufficient in the low-noise regime.

\subsection{Extension to Localized Estimators (PCA)} \label{app:extension}

\begin{corollary}[Sample-wise Error Bound for Local Estimators]
Consider a localized estimator (e.g., PCA Denoiser \cite{Lukoianov2025Locality}) that computes the denoising score independently for each spatial location (or patch) $p$:
\begin{equation}
    \hat{\ff}^{(p)}(\xx_t) = \sum_{i=1}^N w_i^{(p)} \xx_i^{(p)}, \quad w_i^{(p)} \propto \exp\left( -\| \xx_t^{(p)} - \xx_i^{(p)} \|^2 / 2\sigma_t^2 \right).
\end{equation}
Let $S_t$ be a globally selected subset. The approximation error at location $p$ is bounded by:
\begin{equation}
    \| \hat{\ff}^{(p)}_{\cD} - \hat{\ff}^{(p)}_{S_t} \|_2 \le 2 R^{(p)} (N - |S_t|) \exp\left( -\Delta_k^{(p)} \right) + \epsilon_{\text{mismatch}},
\end{equation}
where $\Delta_k^{(p)}$ is the local Logit Gap at position $p$.
\end{corollary}

\noindent \textbf{Remark.}
\thmref{thm:truncation_error} applies \textit{sample-wise} to each spatial location.
The error consists of two parts:
\begin{itemize}
    \item  Truncation Error (The Exponential Term): This follows directly from \thmref{thm:truncation_error}. In the low-noise regime, the local Logit Gap $\Delta_k^{(p)}$ explodes, driving this error to zero. This confirms that \textit{if} the true local neighbors are present in $S_t$, the PCA estimator converges.
    \item Selection Mismatch Error ($\epsilon_{\text{mismatch}}$): This term arises if the globally selected subset $S_t$ fails to include the true local top-$k$ neighbors for position $p$.
    However, as justified in \secref{sec:dynamic_method}, natural images exhibit strong hierarchical consistency.
    A sample that is a ``true neighbor'' at a local patch level typically exhibits high similarity in the global low-frequency proxy used for selection. Furthermore, by maintaining a sufficiently large candidate pool $m_t$, we minimize the probability of exclusion ($\epsilon_{\text{mismatch}} \approx 0$).
\end{itemize}

\section{Additional Experiments}\label{app_sec:exp}

\paragraph{Results of MNIST and Fashion-MNIST.}
We evaluate the performance of \algopt against established analytical baselines across MNIST and Fashion-MNIST.
The results are reported in \tabref{tab:main_comparison_mnist}.
Obviously, our \algopt consistently surpasses the state-of-the-art PCA method \cite{Lukoianov2025Locality} across all MSE and $r^2$ metrics with significant acceleration.
Qualitatively, \figref{fig:main_vis_results} shows that \algopt exhibits superior similarity to the ground-truth neural diffusion outputs (see last row) on these datasets.

\begin{table*}[t]
\centering
\caption{\textbf{Quantitative Comparison of Analytical Denoisers on MNIST and Fashion-MNIST.} 
\emph{Time} (s) and \emph{Memory} (GB) are averaged over 128 samples. 
Best and second-best results are \textbf{bolded} and \underline{underlined}, respectively.}
\label{tab:main_comparison_mnist}
\small
\begin{tabular}{lcccccccc}
    \toprule
    Dataset & \multicolumn{4}{c}{\textbf{MNIST}} & \multicolumn{4}{c}{\textbf{Fashion-MNIST}} \\
    \cmidrule(lr){2-5} \cmidrule(lr){6-9}
    Metric & \multicolumn{2}{c}{Efficacy} & \multicolumn{2}{c}{Efficiency} & \multicolumn{2}{c}{Efficacy} & \multicolumn{2}{c}{Efficiency} \\
    \cmidrule(lr){2-3} \cmidrule(lr){4-5} \cmidrule(lr){6-7} \cmidrule(lr){8-9}
    Method & MSE~($\downarrow$) & $r^2$~($\uparrow$) & Time & Mem. & MSE~($\downarrow$) & $r^2$~($\uparrow$) & Time & Mem. \\
    \midrule
    Optimal \cite{de2022convergence} & 0.047 & 0.435 & 2.806 & 0.013 & 0.033 & 0.514 & 2.608 & 0.013 \\
    Wiener \cite{wiener1949extrapolation} & 0.034 & 0.594 & 0.001 & 0.026 & 0.031 & 0.589 & 0.001 & 0.026 \\
    Kamb \cite{kamb2024analytic} &0.043 & 0.489 & 4.115 &2.188 & 0.051 & 0.294 & 3.903 & 2.188 \\
    \midrule
    PCA \cite{Lukoianov2025Locality} & \underline{0.027} & \underline{0.684} & 2.798 & 0.204 & \underline{0.015} & \underline{0.795} & 2.791 & \underline{0.204} \\
    
    \textbf{\algopt (Ours)} & \textbf{0.020} & \textbf{0.713} & \textbf{0.019} & 0.282 & \textbf{0.011} & \textbf{0.813} & \textbf{0.088} & 0.282\\
    \bottomrule
\end{tabular}
\end{table*}



\end{document}